\pdfoutput=1
\documentclass[11pt,a4paper]{article}
\usepackage[hyperref]{emnlp-ijcnlp-2019}
\usepackage[utf8]{inputenc}
\usepackage{times}
\usepackage{latexsym}
\usepackage{numprint}

\usepackage{url}

\usepackage{soulutf8,amsmath,mathtools,booktabs,tabularx,float,multirow,stfloats}
\usepackage[english, plain]{fancyref}
\usepackage{csquotes}

\definecolorset{cmyk}{GU-}{}{%
  Goethe-Blau,1,0.2,0,0.4;%
  Hellgrau,0.04,0.04,0.05,0.02;%
  Sandgrau,0.12,0.09,0.13,0;%
  Dunkelgrau,0.25,0.25,0.3,0.75;%
  Purple,0.08,1,0.3,0.36;%
  Emo-Rot,0.04,1,0.8,0.07;%
  Senfgelb,0.01,0.25,1,0.05;%
  Gruen,0.62,0.4,0.87,0.09;%
  Magenta,0.08,0.86,0.12,0.12;%
  Orange,0,0.7,1,0.04;%
  Sonnengelb,0,0.12,0.95,0;%
  Helles-Gruen,0.4,0.17,0.81,0.07;%
  Lichtblau,0.8,0,0.06,0.04}
  
\definecolor{HUDarkBlue}{RGB}{16,82,132}
\definecolor{HULightBlue}{RGB}{55,140,195} %% HTML: 378cc3
\definecolor{HUBlueText}{RGB}{0,115,186}
\definecolor{HUDarkGrey}{RGB}{206,207,209}
\definecolor{HUMedGrey}{RGB}{219,220,221}
\definecolor{HULightGrey}{RGB}{241,241,241} %% HTML: f1f1f1
\definecolor{HUDarkOrange}{RGB}{146,76,0} %% HTML: 924c00
\definecolor{LNshadowblue}{RGB}{112,159,203}
\definecolor{LNlightblue}{RGB}{153,196,236}
\definecolor{LNultralightblue}{RGB}{229,240,250}
\definecolor{LNgreen}{RGB}{153,204,51}
\definecolor{GoetheOrange}{RGB}{243,192,57} % Pantone 143
\definecolor{GoetheOrangePlakativ}{RGB}{237,167,45} % Pantone 151
\definecolor{Papier}{RGB}{233,234,192}
\definecolor{agttlightSeminarGrau}{RGB}{219,220,222}
\definecolor{bottomcolor}{rgb}{0.32,0.3,0.38}
\definecolor{middlecolor}{rgb}{0.08,0.08,0.16}
\definecolor{loeweblau}{rgb}{0.33,0.57,0.84}
\definecolor{loewehellblau}{RGB}{219,238,244}
\definecolor{loewegruen}{RGB}{90,162,80}
\definecolor{loeweorange}{RGB}{220,150,55}
\definecolor{DigihumRot}{RGB}{212,0,0}
\definecolor{DigihumBlau}{RGB}{0,94,170}
\definecolor{DigihumHellBlau}{RGB}{221,240,255}
\definecolor{DigihumText}{RGB}{103,102,102}
\definecolor{DigihumHellGrau}{RGB}{214,214,214}
\colorlet{loewegrau}{DigihumText!83!white}
\definecolor{SeminarBlau}{RGB}{0,154,224}
\definecolor{SeminarRot}{rgb}{0.75,0,0}
\definecolor{SeminarGruen}{RGB}{0,128,0}
\definecolor{SeminarOrange}{RGB}{237,167,45} % rgb: {0.93,0.65,0.18}
\definecolor{SeminarGrau}{rgb}{0.32,0.3,0.38} % RGB: {81.6,76.5,96.6}
\colorlet{SeminarHellBlau}{SeminarBlau!25!white}
\colorlet{SeminarHellGruen}{SeminarGruen!25!white}
\colorlet{SeminarHellRot}{SeminarRot!25!white}
\colorlet{SeminarHellOrange}{GoetheOrangePlakativ!25!white}
\colorlet{SeminarHellGrau}{SeminarGrau!35!white}
\colorlet{SeminarSehrHellGrau}{SeminarGrau!15!white}
\colorlet{SeminarSehrSehrHellGrau}{SeminarGrau!5!white}

\colorlet{SeminarBlauGruen}{SeminarBlau!50!SeminarGruen}
\colorlet{SeminarGruenBlau}{SeminarBlau!50!SeminarGruen}
\colorlet{SeminarMix}{SeminarBlau!50!SeminarGruen}
\colorlet{SeminarHellMix}{SeminarMix!25!white}
\colorlet{SeminarHellBlauGruen}{SeminarMix!25!white}
\colorlet{SeminarHellGruenBlau}{SeminarMix!25!white}
\colorlet{SeminarRotOrange}{SeminarRot!50!GoetheOrangePlakativ}
\colorlet{SeminarOrangeRot}{SeminarRot!50!GoetheOrangePlakativ}
\colorlet{SeminarHellOrangeRot}{SeminarOrangeRot!25!white}
\colorlet{SeminarHellRotOrange}{SeminarOrangeRot!25!white}
\colorlet{SeminarRotGruen}{SeminarRot!50!SeminarGruen}
\colorlet{SeminarGruenRot}{SeminarRot!50!SeminarGruen}
\colorlet{SeminarRotBlau}{SeminarRot!50!SeminarBlau}
\colorlet{SeminarBlauRot}{SeminarRot!50!SeminarBlau}
\colorlet{SeminarHellRotBlau}{SeminarRotBlau!25!white}
\colorlet{SeminarHellBlauRot}{SeminarRotBlau!25!white}
\colorlet{SeminarHellRotGruen}{SeminarRotGruen!25!white}
\colorlet{SeminarHellGruenRot}{SeminarRotGruen!25!white}
\colorlet{SeminarOrangeGruen}{SeminarOrange!50!SeminarGruen}
\colorlet{SeminarGruenOrange}{SeminarOrange!50!SeminarGruen}
\colorlet{SeminarHellGruenOrange}{SeminarOrangeGruen!25!white}
\colorlet{SeminarHellOrangeGruen}{SeminarOrangeGruen!25!white}
\colorlet{SeminarOrangeBlau}{SeminarOrange!50!SeminarBlau}
\colorlet{SeminarBlauOrange}{SeminarOrange!50!SeminarBlau}
\colorlet{SeminarHellBlauOrange}{SeminarOrangeBlau!25!white}
\colorlet{SeminarHellOrangeBlau}{SeminarOrangeBlau!25!white}
\colorlet{SeminarSehrHellRot}{SeminarRot!15!white}
\colorlet{SeminarSehrHellGrau}{SeminarGrau!15!white}
\colorlet{SeminarSehrHellGruen}{SeminarGruen!15!white}
\colorlet{SeminarSehrHellBlau}{SeminarBlau!15!white}
\colorlet{SeminarSehrHellOrange}{SeminarHellOrange!15!white}
\colorlet{SeminarOrangeGrau}{SeminarOrange!50!SeminarGrau}
\colorlet{SeminarGrauOrange}{SeminarOrange!50!SeminarGrau}
\colorlet{SeminarRotGrau}{SeminarRot!50!SeminarGrau}
\colorlet{SeminarGrauRot}{SeminarRot!50!SeminarGrau}
\colorlet{SeminarBlauGrau}{SeminarBlau!50!SeminarGrau}
\colorlet{SeminarGrauBlau}{SeminarBlau!50!SeminarGrau}
\colorlet{SeminarGruenGrau}{SeminarGruen!50!SeminarGrau}
\colorlet{SeminarGrauGruen}{SeminarGruen!50!SeminarGrau}
\colorlet{SeminarHellGrauRot}{SeminarGrauRot!25!white}
\colorlet{SeminarHellRotGrau}{SeminarGrauRot!25!white}
\colorlet{SeminarHellGrauBlau}{SeminarGrauBlau!25!white}
\colorlet{SeminarHellBlauGrau}{SeminarGrauBlau!25!white}
\colorlet{SeminarHellGrauGruen}{SeminarGrauGruen!25!white}
\colorlet{SeminarHellGruenGrau}{SeminarGrauGruen!25!white}
\colorlet{SeminarHellGrauOrange}{SeminarGrauOrange!25!white}
\colorlet{SeminarHellOrangeGrau}{SeminarGrauOrange!25!white}

\usepackage{adjustbox,tikz,pgf,pgfplots,etoolbox}
\usetikzlibrary{arrows,arrows.meta,backgrounds,calc,fit,matrix,positioning}
\pgfplotsset{compat=1.14}
\pgfdeclarelayer{background}
\pgfdeclarelayer{foreground}
\pgfsetlayers{background,main,foreground}
\newcounter{dummy}
\def\getNoOfElements#1#2{%
	\setcounter{dummy}{0}% 
	\foreach\dummy in {#1}{\stepcounter{dummy}}%
	\edef#2{\arabic{dummy}}
}

\newcommand*{\LM}[4][xscale=0.35,yscale=0.4]{
	\begin{tikzpicture}[
		font=\small,
		plus/.style={path picture={ 
				\draw[black]
				(path picture bounding box.east) -- (path picture bounding box.west) (path picture bounding box.south) -- (path picture bounding box.north);
		}},
		ArrowRoundSmall/.style={
			rounded corners=.4mm,
		},
		ArrowRoundMedium/.style={
			rounded corners=.1cm,
		},
		cell/.style={
			circle, draw=black, inner sep=0pt, minimum size=0.25 em
		},
		Vert/.style={
			draw=black!80, align=center, minimum height=0.3cm, text width=0.1cm, inner sep=0pt
		},
		letter/.style={
			draw=black, text=black, align=center, minimum height=1.5 cm, text width=1 cm, inner sep=0pt
		},
		Label/.style={
			draw=black, anchor=center, align=center, text centered, text height=7pt, font=\footnotesize
		},
		LabelSmall/.style={
			anchor=west, font=\scriptsize, align=center, text centered
		},
		x=1cm,
		y=-1cm
		]
		\tikzset{>={Latex[length=0.75mm]}}
		\edef\cells{#2}
		\getNoOfElements{#4}\words
		\pgfmathparse{\words-1}
		\edef\wmo{\pgfmathresult}
		
		\begin{scope}[#1]
			\begin{pgfonlayer}{main}
				\foreach \name / \x in {1,...,\cells}
				    \node [cell] (F-\name) at (\x,1) {};
				\foreach \name / \x in {1,...,\cells}
				    \node [cell] (B-\name) at (\x+0.5,1.5) {};
				\foreach \name / \x in {1,...,\cells}{
					\draw [ArrowRoundSmall] (\x+0.25,0) -| (\x+0.25,0.3) [->] -| (F-\name.north);
					\draw [ArrowRoundSmall] (\x+0.25,0) -| (\x+0.25,0.3) [->] -| (B-\name.north);
					\draw [ArrowRoundSmall] (B-\name.south) |- (\x+0.25,2) [->] |- (\x+0.25,2.5);
					\draw [ArrowRoundSmall] (F-\name.south) |- (\x+0.25,2) [->] |- (\x+0.25,2.5);
				}
			
				\draw [decorate,decoration={brace,amplitude=0.15cm},xshift=-0.2cm,yshift=0cm] (0,-0.5) -- (0,-4.5) node [midway, rotate=90, anchor=south, yshift=0.1cm] (Label-Input) {Input};
				\node [draw=none] at ($(\cells+1, 0) - (Label-Input.north)$) (Spacer) {};
				\foreach \char [count=\x from 1] in {#3} {
					\node [rectangle, anchor=south, draw=black, inner sep=0, minimum height=1 cm, text width=1 cm, #1, align=center] (C-\x) at (\x+0.25,-0.5) {};
					\node [anchor=south, yshift=-0.075cm] (CL-\x) at (C-\x.south) {\char};
					\draw [ArrowRoundSmall] (C-\x.south) [->] -- (\x+0.25,-0.1);
				}
				
				\pgfmathparse{\cells+1.5}
				\edef\cpo{\pgfmathresult}
				\pgfmathparse{\cells-1}
				\edef\cmo{\pgfmathresult}
				
				\draw [->] (F-\cells) -- ++(1, 0);
				\draw [<-] (B-\cells) -- ++(0.5, 0);
				\draw [->] (B-1) -- ++(-1, 0);
				\draw [<-] (F-1) -- ++(-0.5, 0);
				
				\foreach \name [count=\next from 2] in {1,...,\cmo}{
					\draw [->] (F-\name.east) -- (F-\next.west);
					\draw [->] (B-\next.west) -- (B-\name.east);
				}
			
				\foreach \a / \b / \str [count=\id from 1] in {#4} {
					\pgfmathparse{\a+0.25}
					\edef\A{\pgfmathresult}
					\pgfmathparse{\b+0.25}
					\edef\B{\pgfmathresult}
					\pgfmathparse{(\B+\A)/2}
					\edef\center{\pgfmathresult}
					\node[rectangle, anchor=center, draw=black](R-\id) at (\center,4) {$\mathrm{r}_\text{\str}$};
					\draw [ArrowRoundMedium] (\A,2.55) |- ([yshift=0.5cm]R-\id.north) [->] |- (R-\id.north);
					\draw [ArrowRoundMedium] (\B,2.55) |- ([yshift=0.5cm]R-\id.north) [->] |- (R-\id.north);
				}
				
				\foreach \a / \b / \str / \lbl / \syl [count=\id from 1] in {#4} {
					\matrix [
					matrix of nodes,
					nodes=Vert,
					column sep=0cm,
					below=0.2cm of R-\id,
					ampersand replacement=\&,
					nodes={fill=SeminarBlau}
					] (ft-\id) {
						{} \& {} \& {} \& {} \& {} \& {} \& {} \& {} \& {} \& {} \& \\
					};
					
					\matrix [
					matrix of nodes,
					nodes=Vert,
					column sep=0cm,
					below=-0.15cm of ft-\id,
					ampersand replacement=\&,
					nodes={fill=SeminarBlau}
					] (w2v-\id) {
						{} \& {} \& {} \& {} \& {} \& {} \& {} \& {} \& {} \& {} \& \\
					};
					
					\matrix [
					matrix of nodes,
					nodes=Vert,
					column sep=0cm,
					below=0.0cm of w2v-\id,
					ampersand replacement=\&,
					nodes={fill=SeminarRot}
					] (BPEmb-\id) {
						{} \& {} \& {} \& {} \& {} \& {} \& {} \& {} \& {} \& {} \& \\
					};
					
					\node [below=0.1cm of BPEmb-\id,draw,circle,plus,inner sep=0pt,minimum size=0.35 cm](stack-\id) {};
					
					\draw 	let 	\p{RF} = ($ (R-\id.west) - (BPEmb-\id.west) $),
									\n{diff} = {max(\x{RF}+1, 10)},
									\p{PCE} = (R-\id.center),
									\p{ft} = (ft-\id.center),
									\p{w2v} = (w2v-\id.center),
									\p{BPEmb} = (BPEmb-\id.center),
									\p{intersec} = ($ (R-\id.west) - (\n{diff}, 0) $)
							in
								(ft-\id-1-1.west) -| (\x{intersec}, \y{ft})
								(w2v-\id-1-1.west) -| (\x{intersec}, \y{w2v}) 
								(BPEmb-\id-1-1.west) -| (\x{intersec}, \y{BPEmb})
								(R-\id.west) -| (\x{intersec}, \y{PCE})
								[-{Latex[length=1.25mm]}] |- (stack-\id.west);
					
					\path (ft-\id.south) |- (0,-3.7) node [midway, anchor=center, draw=black, fill=SeminarHellBlau!50!SeminarBlau, inner sep=2pt] (ft-\id-str) {$T(\text{\str})$};
					\path (BPEmb-\id.south) |- (0,-2.3) node [midway, anchor=center, draw=black, fill=SeminarHellRot!50!SeminarRot, inner sep=2pt] (ft-\id-str) {$S(\text{\syl})$}; 	
				}
			
				\foreach \id in {1,...,\words} {
					\node [cell] at ($ (stack-\id.south) + (-0.25,1) $) (BiLSTM-F-\id) {};
					\node [cell] at ($ (stack-\id.south) + (0.25,1.5) $) (BiLSTM-B-\id) {};
					\draw [ArrowRoundSmall] (stack-\id.south) |- ($ (stack-\id.south) + (-0.25,0.5) $) [->] -- (BiLSTM-F-\id.north);
					\draw [ArrowRoundSmall] (stack-\id.south) |- ($ (stack-\id.south) + (0.25,0.5) $) [->] -- (BiLSTM-B-\id.north);
					\draw [ArrowRoundSmall] (BiLSTM-F-\id.south) -- ($ (stack-\id.south) + (-0.25,2) $) [->] -| ($ (stack-\id.south) + (0,2.5)$);
					\draw [ArrowRoundSmall] (BiLSTM-B-\id.south) -- ($ (stack-\id.south) + (0.25,2) $) [->] -| ($ (stack-\id.south) + (0,2.5)$);
				}
			
				\draw [ArrowRoundSmall] (BiLSTM-F-\words.east) [->] -- ++(1,0);
				\draw [ArrowRoundSmall] (BiLSTM-B-\words.east) [<-] -- ++(0.5,0);
				\draw [ArrowRoundSmall] (BiLSTM-F-1.west) [<-] -- ++(-0.5,0);
				\draw [ArrowRoundSmall] (BiLSTM-B-1.west) [->] -- ++(-1,0);
				\foreach \id [count=\next from 2] in {1,...,\wmo} {
					\draw [ArrowRoundSmall] (BiLSTM-F-\id.east) [->] -- (BiLSTM-F-\next.west);
					\draw [ArrowRoundSmall] (BiLSTM-B-\id.east) [<-] -- (BiLSTM-B-\next.west);
				}
			
				\foreach \id in {1,...,\words} {
					\node[rectangle, anchor=north, draw=black] (CRF-\id) at ($ (stack-\id.south) + (0,2.5)$) {};
				}
				\foreach \id [count=\next from 2] in {1,...,\wmo} {
					\draw [ArrowRoundSmall] (CRF-\id.east) [<->] -- (CRF-\next.west);
				}
			
				\foreach \a / \b / \str / \lbl [count=\id from 1] in {#4} {
					\node[rectangle, anchor=north, draw=black] (Label-\id) at ($ (CRF-\id.south) + (0,1) $) {\lbl};
					\draw [ArrowRoundSmall] (CRF-\id.south) [->] -- (Label-\id.north);
				}
	        \end{pgfonlayer}
				
			\begin{pgfonlayer}{background}
				\draw [draw=none, fill=SeminarSehrHellGrau]  (\cells+1.5, 2.5) rectangle ( ft-1.north -| 0,0 );
				\draw [draw=black, -latex, fill=SeminarHellOrange!50] (0, -0.1) rectangle (\cells+1.5, 2.5);
				\fill [fill=SeminarSehrHellBlau] ( ft-1.north -| 0,0 ) rectangle ( w2v-1.south -| \cells+1.5,0 );
				\fill [fill=SeminarSehrHellRot] ( BPEmb-1.north -| 0,0 ) rectangle ( BPEmb-\words.south -| \cells+1.5, 0);
				\node [draw=none] at ($ (stack-\words.south) + (0,0.25) $) (Y2) {};
				\path (R-1.west) -| (0,0) coordinate [midway] (R-Center);
				\path ( ft-1.north -| 0,0 ) -- ( ft-1.south -| 0,0 ) coordinate [midway] (ft-Center);
				\path ( w2v-1.north -| 0,0 ) -- ( w2v-1.south -| 0,0 ) coordinate [midway] (w2v-Center);
				\path ( BPEmb-1.north -| 0,0 ) -- ( BPEmb-\words.south -| 0, 0) coordinate [midway] (BPEmb-Center);
				\path ( BPEmb-1.south -| 0,0 ) -- ( Y2 -| 0,0 ) coordinate [midway] (Stacking-Center);
				\path ($ (Y2) + (0,1) $) -| (0,0) coordinate [midway] (BiLSTM-Center);
				\path (CRF-1) -| (0,0) coordinate [midway] (CRF-Center);
				
				\node [draw=none] at ($ (stack-\words.south) + (0,2.25) $) (Y1) {};
				\fill [fill=SeminarSehrHellGruen] ( Y1 -| 0,0 ) rectangle ( Y2 -| \cells+1.5,0 );
				\node [draw=none] at ($ (CRF-\words.south) + (0,0.25) $) (Y2) {};
				\fill [fill=SeminarHellMix!50!SeminarSehrHellGruen] ( Y1 -| 0,0 ) rectangle ( Y2 -| \cells+1.5,0 );
				\node [draw=none] at ($ (stack-\words.south) + (0,0.25) $) (Y1) {};
				\draw [draw=black] ( Y1 -| 0,0 ) rectangle ( Y2 -| \cells+1.5,0 );
			\end{pgfonlayer}
			
			\begin{pgfonlayer}{foreground}
				\path (0, -0.1) -- (\cells+1.5, 2.5) node (CSE-Label) [Label, fill=SeminarHellOrange, midway] 
				{Character Language Model};
				\node [LabelSmall, label=center:] (R-Label) at (R-Center)  {PCE};
				\node [LabelSmall, label=center:] (ft-Label) at (ft-Center)  {fastText};
				\node [LabelSmall, label=center:] (w2v-Label) at (w2v-Center)  {wang2vec};
				\node [LabelSmall, label=center:] (BPEmb-Label) at (BPEmb-Center)  {BPEmb};
				\node [LabelSmall, label=center:] (Stacking-Label) at (Stacking-Center)  {Stacking};
				\node [LabelSmall, label=center:] (BiLSTM-Label) at (BiLSTM-Center)  {BiLSTM};
				\node [LabelSmall, label=center:] (CRF-Label) at (CRF-Center)  {CRF};
				\path (Y1 -| 0,0) -- (Y2 -| \cells+1.5,0) node (BiLSTM-CRF-Label) [Label, fill=SeminarHellMix!50!SeminarSehrHellGruen, midway, yshift=-0.05cm] {Sequence Labeling Model};
			\end{pgfonlayer}
		\end{scope}
	\end{tikzpicture}
}

\newcommand{\charExample}{%
        \begin{tikzpicture}[inner sep=2pt, baseline=(base)]%
        \node [draw=black] at (-2mm,0){C};
        \node (base) at (0,-0.8ex) {};
        \end{tikzpicture}%
    }

\newcommand{\tokenExample}{%
        \begin{tikzpicture}[inner sep=2pt, baseline=(base)]%
        \node [draw=black,fill=SeminarHellBlau!50!SeminarBlau] at (-2mm,0){$T$};
        \node (base) at (0,-0.8ex) {};
        \end{tikzpicture}%
    }

\newcommand{\syllableExample}{%
        \begin{tikzpicture}[inner sep=2pt, baseline=(base)]%
        \node [draw=black,fill=SeminarHellRot!50!SeminarRot] at (-2mm,0){$S$};
        \node (base) at (0,-0.8ex) {};
        \end{tikzpicture}%
    }

\aclfinalcopy

\title{When Specialization Helps: Using Pooled Contextualized Embeddings to Detect Chemical and Biomedical Entities in Spanish}

\author{Manuel Stoeckel \\
	Goethe University Frankfurt \\
	Text Technology Lab \\
	{\tt manuel.stoeckel@stud.uni-frankfurt.de} \\\AND
	Wahed Hemati \\
	Goethe University Frankfurt \\
	Text Technology Lab \\
	{\tt hemati@em.uni-frankfurt.de} \\\And
	Alexander Mehler \\
	Goethe University Frankfurt \\
	Text Technology Lab \\
	{\tt mehler@em.uni-frankfurt.de} \\}

\date{\today}

\begin{document}
	\maketitle
	\begin{abstract}
		The recognition of pharmacological substances, compounds and proteins is an essential preliminary work for the recognition of relations between chemicals and other biomedically relevant units.
		In this paper, we describe an approach to Task 1 of the PharmaCoNER Challenge, which involves the recognition of mentions of chemicals and drugs in Spanish medical texts.
		We train a state-of-the-art BiLSTM-CRF sequence tagger with stacked Pooled Contextualized Embeddings, word and sub-word embeddings using the open-source framework FLAIR.
		We present a new corpus composed of articles and papers from Spanish health science journals, termed the \textit{Spanish Health Corpus}, and use it to train domain-specific embeddings which we incorporate in our model training.
		We achieve a result of $89.76\%$ F1-score using pre-trained embeddings and are able to improve these results to $90.52\%$ F1-score using specialized embeddings.
	\end{abstract}

	\section{Introduction}
    	Efficient access to information on chemicals and pharmaceutical units has become increasingly important for researchers in various chemical disciplines.
    	However, manual annotation of these units to create knowledge bases is a laborious process given the ever-increasing number of papers and patents in bio/chemical and pharmaceutical research.
    	Thus, \textit{Natural Language Processing} (NLP) can be employed to detect such entities and their relations from the relevant literature.
    	Previous work has been successful in detecting and classifying chemical substances or in extracting complex relations between chemical substances \cite{Krallinger:2015:CHEMDNER,Hemati:Mehler:2019a}.

    	While most NLP research is conducted on English datasets, there are a considerable number of non-English biomedically relevant texts written in other languages, e.g.\ clinical texts.
    	In order to advance the further development of biomedical and pharmaceutical entity recognition facing this linguistic diversity, the PharmaCoNER task challenges participants with \textit{Named Entity Recognition} (NER) for pharmacological substances, compounds and proteins on a Spanish corpus \citep{GonzalezAgirre:2019:BioNLP:PharmaCoNER}.
    	The PharmaCoNER task belongs to the \textit{BioNLP Open Shared Tasks 2019} (BioNLP-OST 2019) Workshop and distinguishes two tracks: 
    	the first track focuses on NER offset and entity classification, while the second task deals with concept indexing.

    	In this paper we present an architecture for NER of chemical and pharmacological units in Spanish texts that produces an F-score of up to 90\%.
    	Source code and instructions for reproducing these results are available on GitHub%
        \footnote{\url{www.git.io/JenqE}} 
        % \footnote{\url{www.github.com/texttechnologylab/EsPharmaNER}}
        and we are offering an interactive web service for testing our models.%
        \footnote{\url{espharmaner.texttechnologylab.org}}
    	The article is organized as follows: First, we describe the resources used to train our model and explain our methodical approach.
    	This includes a detailed description of the PharmaCoNER dataset and the kind of preprocessing we performed on the input texts. 
    	Afterwards, we give a thorough description of our architecture.
        Finally, we discuss our results and give our conclusions.

	\section{Materials and Methods}
    	\subsection{Datasets}
        	In this section, we describe the datasets used in our experiments and the architecture of our NER tagger.
        	
        	\paragraph{PharmaCoNER}
            	The corpus accompanying the PharmaCoNER task, that is, the \textit{Spanish Clinical Case Corpus} (SPACCC), contains \numprint{1000} manually classified clinical cases and comprises \numprint{396988} token \cite{GonzalezAgirre:2019:PharmaCoNER:Corpus}.
            	The corpus was derived from open access Spanish medical publications and (according to the creators) shows properties of both biomedical and medical literature as well as clinical records.

            	The SPACCC corpus is given in brat standoff format\footnote{\url{brat.nlplab.org/standoff.html}} as two separate files per document, one containing the plain text, the other containing the annotations with character level offsets on the raw text.
            	We converted the corpus into a CoNLL2003 compatible format, applying common whitespace tokenization and splitting tokens on non-alphanumeric characters, as this increased the performance of our model.

        	\paragraph{Spanish Health Corpus}
            	In this section, we describe the Spanish Health Corpus, a collection of 7353 diverse Spanish health and science journal articles and papers.
            	The corpus was obtained from SciELO\footnote{\url{www.scielo.org}} by means of an automated crawler.%
            	\footnote{See our GitHub repository for the list of documents.}
            	The content of the articles in this corpus was downloaded as embedded text from the respective websites and stripped of any structural elements, like HTML tags.
            	Then, the raw text was split into sentences using \textsc{deep-eos}, a neural network sentence boundary detection tool created by Stefan Schweter which is publicly available on GitHub.\footnote{\url{www.github.com/stefan-it/deep-eos}}

            	We trained a Spanish \textsc{deep-eos} LSTM model on \numprint{100000} Spanish Wikipedia sentences extracted from the Leipzig Corpora Collection \cite{Goldhahn:2012:LeipzigCorpora}.
            	Our \textsc{deep-eos} model achieves an accuracy of 99.65\% on separate \numprint{100000} test sentences.
            	The resulting sentences were then tokenized based on the procedure mentioned in the previous section.
            	This resulted  in a set of \numprint{957648} sentences containing \numprint{32346137} words in total.
            	We used this corpus to train special word embeddings for our system that we believe have a positive impact on the performance of our models.

    	\subsection{System Architecture}
        	Our system was built with FLAIR \citep{Akbik:etal:2019:ACL:FLAIR}, an easy to use open-source NLP framework that is able to produce state-of-the-art results for sequence tagging tasks \citep[eg.][]{Akbik:2018:COLING:CSE,Akbik:2019:NAACL:PooledCSE}.
        	We follow the approach of \citeauthor{Akbik:etal:2019:ACL:FLAIR}, using FLAIR to stack (i.e. concatenate) character and word embeddings to improve recognition rates.
        	We further expand this model by adding sub-word embeddings to the stacked embeddings.
        	These stacked embeddings serves as input for a BiLSTM-CRF sequence tagger \citep{Huang:2015:Bidirectional}.
        	
        	\begin{figure*}[ht!]
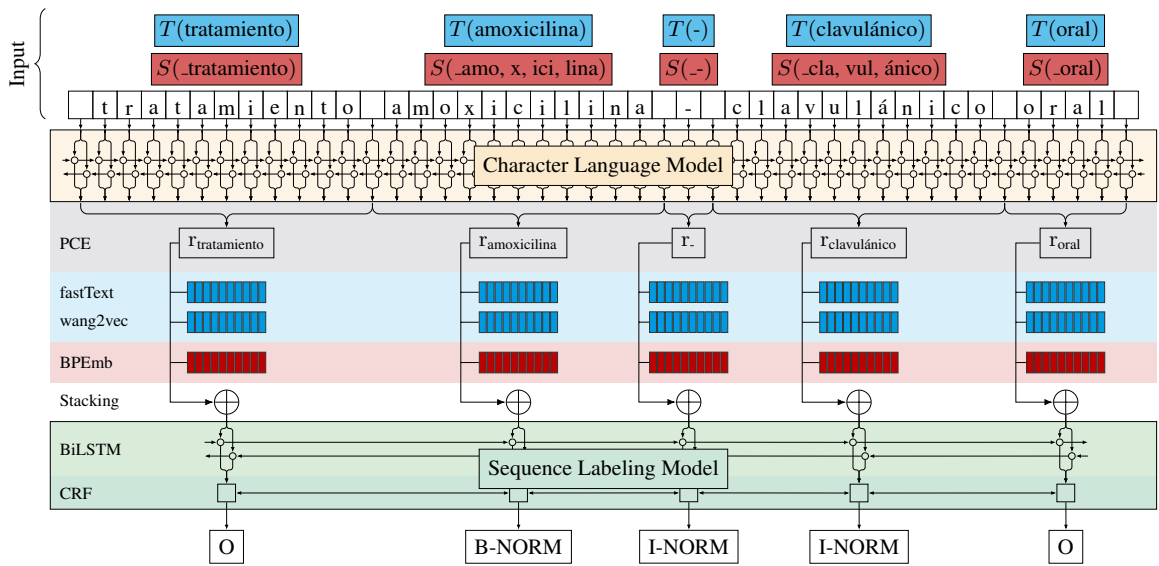

        		\centering
        		\begin{adjustbox}{max width=\textwidth}
        			\LM{44}{,t,r,a,t,a,m,i,e,n,t,o, ,a,m,o,x,i,c,i,l,i,n,a, ,-, ,c,l,a,v,u,l,\'{a},n,i,c,o, ,o,r,a,l,}{{1/13/tratamiento/O/\_tratamiento},{13/25/amoxicilina/B-NORM/\_amo, x, ici, lina},{25/27/-/I-NORM/\_-},{27/39/clavul\'{a}nico/I-NORM/\_cla, vul, \'{a}nico},{39/44/oral/O/\_oral}}
        		\end{adjustbox}
        		\caption{
            		The architecture of the best performing model in our experiments.
            		The PCEs are generated from \charExample character features,
            		while \textsc{fastText} and \textsc{wang2vec} embeddings are trained on \tokenExample tokens,
            		and \textsc{BPEmb} uses \syllableExample syllable input. 
            		The embeddings are stacked and serve as input for a BiLSTM-CRF Sequence Labeling Model.
        		}
        		\label{fig:architecture}
        	\end{figure*} 
        	
        	For our best performing model, we used two different token-level embeddings, a \textsc{wang2vec}-based embedding \citep{Ling:2015:NAACL} and a \textsc{fastText}-based embedding \citep{Bojanowski:2017:TACL:FastText}, a single byte-pair sub-word embedding \citep{Heinzerling:2018:LREC:BPEmb} and one context sensitive character-level language model \citep{Akbik:2019:NAACL:PooledCSE}.
        	\Fref{fig:architecture} gives a visual depiction of our best performing model.
        	The following paragraphs describe the used embeddings in more detail.
        	
        	\paragraph{Pooled Contextualized Embeddings}
            	\textit{Contextualized String Embedding}s \citep[CSEs]{Akbik:2018:COLING:CSE} use pre-trained character-level language models from which hidden states at the start and end character positions of each word are extracted to create embeddings for any string in sentence contexts.
            	This model is further developed by \citet{Akbik:2019:NAACL:PooledCSE} who introduce an expansion to CSEs in terms of \textit{Pooled Contextualized Embeddings} (PCEs).

            	PCEs tackle the problem of embedding rare words by applying a pooling operation on different contextual embeddings of the word.
            	The authors follow the idea that words which occur in under-specified contexts should be familiar to the reader from previous mentions.
            	So when a word is processed during the training of a character-level language model, all previous contextualized instances of the word are pooled and concatenated with the current instance to create a \enquote{global} word representation \citep{Akbik:2019:NAACL:PooledCSE}.
            	The authors experiment with three pooling operations (\textit{min}, \textit{max} and \textit{mean}).
            	In this way, they are able to achieve state-of-the-art results in four major NER tasks \citep{Akbik:2019:NAACL:PooledCSE}.
            	
            	In our architecture, we employ pre-trained Spanish Pooled Contextualized Embeddings.\footnote{These models were trained by Yihwa Kim\newline (\url{www.github.com/iamyihwa}).}

        	\paragraph{wang2vec Embeddings}
            	This model, proposed by \citet{Ling:2015:NAACL}, is an extension of the token-level \textsc{word2vec} model of \citet{Mikolov:2013:NAACL:Word2Vec}.
            	During training, \textsc{wang2vec} makes a prediction for each neighboring position of the target word instead of making a single prediction for all neighbours.
            	Thus, the resulting embeddings are better at capturing syntactic, positional information \cite{Ling:2015:NAACL}.

            	We trained 300 dimensional \textsc{wang2vec}-based embeddings based on 100 iterations using default parameters on the Spanish Health Corpus.

        	\paragraph{fastText Embeddings}
            	Unlike \textsc{word2vec} or \textsc{wang2vec}, \textsc{fastText} \citep{Bojanowski:2017:TACL:FastText} models words as sets of character n-grams, where all n-grams from sizes 3-6 are used during training.
            	\textsc{fastText} can thus represent rare words that were not present in the vocabulary of the training files if their skip-grams were observed during training.
            	Before the words are split into n-grams, special boundary symbols are added.
            	The embeddings are thus also able to learn information about word prefixes and suffixes \citep{Bojanowski:2017:TACL:FastText}.
            	We used pre-trained 300 dimensional Spanish \textsc{fastText} embeddings from \citet{Grave:2018:LREC:FastTextEmbeddings} in our initial submission to the PharmaCoNER task.\footnote{\url{www.fasttext.cc/docs/en/crawl-vectors.html}}

            	We replaced them with our own 300 dimensional embeddings trained on the Spanish Health Corpus with standard parameter settings during our experimental phase.

        	\paragraph{Byte-Pair Embeddings.}
            	Similar to \textsc{fastText}, Byte-Pair embeddings \citep[\textsc{BPEmb}]{Heinzerling:2018:LREC:BPEmb} are trained on a pre-processed corpus that contains sub-word entities.
            	But in contrast to \textsc{fastText}, words in the training corpus are represented as combinations of \textit{syllables} instead of skip-grams.
            	These syllables or \textit{subword units} are learned from the corpus prior to the segmentation using Byte-Pair-Encoding \citep{Sennrich:2016:ACL:BytePairEncoding} for a predefined number.

            	In our experiments, we used pre-trained 300 dimensional Spanish Byte-Pair embeddings made available by \citet{Heinzerling:2018:LREC:BPEmb} with a syllable vocabulary size of \numprint{100000}.\footnote{\url{www.github.com/bheinzerling/bpemb}}

    	\subsection{Experiments}
        	We conducted extensive experiments to optimize our models.
        	The ease of use of FLAIR enables us to swap embeddings and optimizers on the fly and perform a state-of-the-art hyper-parameter search.
        	Following the \enquote{best known configurations} for NER tasks in English, German and Dutch according \citeauthor{Akbik:etal:2019:ACL:FLAIR}'s GitHub repository,\footnote{\url{www.github.com/zalandoresearch/flair}} we trained the BiLSTM-CRF sequence tagger with a hidden size of 256, a single LSTM layer (unless stated otherwise) and no dropout.
        	We used common \textit{Stochastic Gradient Descent} (SDG) with a learning rate of 0.1, mini-batch size of 32, an annealing rate of 0.5 with a patience of 3 and default parameters otherwise.
        	The training takes about 80 epochs with these settings.

        	We performed a parameter search with FLAIR's wrapper of the hyper parameter selection tool \textsc{Hyperopt} \citep{Bergstra:2013:MSM:3042817.3042832}.
        	We chose our initial search parameters similar to the search conducted by \citet{Akbik:2019:NAACL:PooledCSE}, which includes a learning rate $\in \left\{0.01, 0.05, 0.1\right\}$ and mini-batch size $\in \left\{8, 16, 32\right\}$.
        	Using this parameter set we were unable to improve our models performance over the performance using the suggested ones.
        	In addition, we ran a sparse parameter search with a different array of possible choices: hidden size $\in \left\{256, 512\right\}$, dropout $\in \left[0,0.5\right]$,
        	number of RNN layers $\in \left\{1, 2\right\}$ and learning rate $\in \left\{ 0.05, 0.1, 0.15\right\}$.
        	While all of the trained models performed very well, we were unable to outperform our previous best model.

        	All experiments were performed either on a NVIDIA GTX 1660 with 6 GiB VRAM available or on a NVIDIA GTX 1080 Ti with 11 GiB VRAM available.

	\section{Evaluation}
    	\paragraph{Results}
        	\Fref{tab:scores-PharmaCoNER} compares the scores of our systems.
        	All scores were computed using the official evaluation script provided by the organizers of the PharmaCoNER task on the gold standard test data, which was released after the end of the challenge phase.
        	After establishing a baseline using mean-pooled PCEs only, we added pre-trained Byte-Pair embeddings (\textsc{BPEmb-pre}) and pre-trained \textsc{fastText} (\textsc{FT-pre}) embeddings.
        	While Byte-Pair embeddings alone were able to increase the performance of the model by $+3.61\%$ F1-score, further adding pre-trained \textsc{fastText} embeddings only increased the systems performance about $+0.11\%$ for a total of $+3.72\%$ against our baseline.
        	This confirms the observations of \citet{Akbik:2019:NAACL:PooledCSE} according to which stacking token-level embeddings on PCEs can improve the performance of the model significantly.
        	Adding a second LSTM layer to the BiLSTM sequence tagger decreased the models F1-score by $0.54\%$ as can be seen in the second entry in row 3 of \fref{tab:scores-PharmaCoNER}.

        	\begin{table}[ht!]
        		\centering
        		\small
        		\begin{tabularx}{\linewidth}{X>{\ttfamily}c>{\ttfamily}c>{\ttfamily}c}
        			\toprule
        			Model & \normalfont{F1-Score} & \normalfont{Precision} & \normalfont{Recall} \\
        			\midrule
        			PCE-\textsc{pre} & 86.04 & 88.59 & 83.64 \\
        			(BSE) &&&\\
        			\midrule
        			BSE + \textsc{BPEmb-pre}$^\dag$ & 89.65 & 90.45 & 88.86 \\
        			(SBM) &&&\\
        			\midrule
        			SBM + \textsc{FT-pre} &&&\\
        			~ 1 LSTM layer$^{\dag\ddag}$         & 89.76 & 90.69 & 88.85 \\
        			~ 2 LSTM layers$^\dag$              & 89.22 & 89.10 & 89.34 \\
        			\midrule
        			SBM + \textsc{FT$^S$} + \textsc{w2v$^S$} &&&\\
                    ~ min-pooled$^\ast$  & 90.31          & 90.02          & 89.71 \\
        			~ max-pooled$^\ast$  & 90.34          & \textbf{90.97} & 89.71 \\
        			~ mean-pooled$^\ast$ & \textbf{90.52} & 90.79          & \textbf{90.30} \\
        			\bottomrule
        		\end{tabularx}
        		\caption{%
            		All scores in \%.
            		BSE denotes our baseline, while SBM denotes our first submission model.
            		The notation \enquote{X + Y} is to be read as \enquote{X stacked with Y}.
            		Legend: $^\dag$ indicates challenge submissions,\newline
            		$^\ddag$ indicates the best challenge submission,\newline
            		$^S$ indicates self-trained specialized embeddings,\newline
            		$^\ast$ indicates models built after the challenge deadline.
        		}
        		\label{tab:scores-PharmaCoNER}
        	    \vspace*{-2ex}
        	\end{table}
        	
        	After the challenge phase, we replaced the pre-trained \textsc{fastText} embeddings with our self-trained, specialized embeddings (\textsc{FT}$^S$) and added the specialized \textsc{wang2vec} (\textsc{w2v}$^S$) embeddings.
        	This increased the performance of the system to $90.34\%$ F1-score.
        	The choice of mean-pooled PCEs in favor of min-pooled PCEs resulted in a further increase in performance to $90.52\%$ F1-score, representing a total increase of $+4.48\%$ over our baseline and $+0.76\%$ over our best result during the challenge phase, while choosing max-pooled PCEs results in the highest precision score of all our models ($90.97\%$).

	\section{Conclusion and Future Work}
    	Our experiments show that with current frameworks like FLAIR it is possible to achieve very good test results with little time spent on system development or implementation.
    	Good results can be achieved with pre-trained models and embeddings that are available in many languages thanks to the NLP community's ongoing efforts.

    	Our experiments confirm our expectations regarding the usability of special embeddings.
    	The embeddings that are trained on the Spanish Health Corpus contribute to significantly increasing the performance of our system, even with such a small training corpus.
        Our results show that the use of domain-specific embeddings can significantly improve the performance of sequence tagging models even in the case of small corpora.

    	We will be continuing our experiments in due time, using larger corpora for our training.
    	In the mean time all our results, datasets and code necessary to reproduce our experiments have been made publicly available on GitHub and can be tested with an interactive web service.

	\section*{Acknowledgements}
	    We would like to thank the anonymous reviewers for their fair opinions and the organisers for their patience and help with problems during the submission of the workshop papers.

\bibliography{PharmaCoNER_2019.bib}
\bibliographystyle{acl_natbib}

\end{document}